# Deep Learning-Based Cyber-Attack Detection Model for Smart Grids

[1]Mojtaba Mohammadi, Arshia Aflaki, Abdollah Kavousifard, *Senior Member, IEEE,* Mohsen Gitizadeh

*Abstract*—In this paper, a novel artificial intelligence-based cyber-attack detection model for smart grids is developed to stop data integrity cyber-attacks (DIAs) on the received load data by supervisory control and data acquisition (SCADA). In the proposed model, first the load data is forecasted using a regression model and after processing stage, the processed data is clustered using the unsupervised learning method. In this work, in order to achieve the best performance, three load forecasting methods (i.e. extra tree regression (ETR), long short-term memory (LSTM) and bidirectional long short-term memory (BiLSTM)) are utilized as regression models and their performance is compared. For clustering and outlying detection, the covariance elliptic envelope (EE) is employed as an unsupervised learning method. To examine the proposed model, the hourly load data of the power company of the city of Johor in Malaysia is employed and Two common DIAs, which are DIAs targeting economic loss and DIAs targeting blackouts, are used to evaluate the accuracy of detection methods in several scenarios. The simulation results show that the proposed EE-BiLSTM method can perform more robust and accurate compared to the other two methods.

## I. Introduction

Generally, a microgrid (MG) is defined as a low voltage self-governing electric system that mainly includes dispatchable loads, distributed generation (DG) units, energy storage devices, etc. The concept of MG brings several benefits to electric grids. For instance, it can facilitate the implementation of renewable sources, improve reliability, reduce operation costs, decrease greenhouse emissions, etc. [1, 2]. According to [4], the installed MG capacity in the United States (US) is increasing rapidly in recent years. In this regard, various aspects of these systems have become a hot research topic among researchers recently. MGs are considered complicated cyber-physical systems where the physical layer mainly includes legacy devices such as DGs, loads, protection systems, etc. The cyber layer includes communication and data-based devices such as smart metering devices, data aggregators, etc. Generally, in MGs, the advanced metering infrastructure (AMI) is the key technology that provides bidirectional communication between different system parts. Since MGs' intelligence and reliable operation are tied with real-time monitoring of the system, cybersecurity in these systems is a vital issue that must be taken into consideration. Widespread implementation of MGs in modern electric grids and high penetration of data-based platforms and devices in MGs has made these systems a great target for adversaries to cause damage to societies. According to a recent report from the US Department of Homeland Security, hackers attacked local electric power companies more than 200 times during 2013 and 2014 [4]. Therefore, it is expected to see a growing concern on cybersecurity in smart grids. In [3], authors developed an intrusion detection system to detect identity-based cyber-attacks in wireless-based AMIs. In that paper, the authors used the sequential hypothesis testing method along with the received signal strength of meters' antennas to distinguish different signal sources and detect cyber-attacks. An evolutionary machine learning-based cyber-attack detection model is proposed in [4]. In that paper, a novel method based on neural networks, prediction intervals, and symbiotic organism search is developed to detect DIAs on smart meters within MGs. Authors in [5] used the LSTM and prediction intervals and proposed an intrusion detection system for false data injection attacks (FDIAs) on load demand in microgrids. In [6], a secured framework for the optimal operation of HMGs is proposed, and the directed acyclic graph was utilized as an intrusion prevention system to improve data communication in the system. A denial of service attack-resilient framework for the optimal scheduling and operation of IoT-based MGs is proposed in [7]. Reference [8] investigated false data injection attacks in smart AC islands and proposed a cyber-attack detection model based on wavelet singular entropy to detect FDIAs on state estimation. A distributed cyber-attack detection method for DC MGs is proposed in [9]. The proposed methodology in that paper uses the Luenberger observer along with a bank of unknown input observers at each subsystem to detect attacks. The overall cyber vulnerabilities of smart grids control systems and various cyber-attack scenarios on these systems are investigated in [10]. In [11], authors used generative adversarial networks and the firefly algorithm and proposed an evolutionary deep learning-based cyber-attack detection model for securing data communication in electric vehicles. The proposed methodology in that paper classifies message frames transferring between ECU and other devices in electric vehicles. In [12], a secured framework based on blockchain-enabled IoT is proposed to improve data security within network MGs. The proposed methodology in that paper tested on a practical test case containing residential, commercial, and crucial MGs. In [13], a Machin learning-based cyber-attack detection model based on nonlinear auto-regressive exogenous neural networks model and time series analysis was proposed to enhance data security in DC MGs. A comprehensive survey on cybersecurity in MGs is presented in [14]. That paper first discussed the structure of MGs as a cyber-physical system and then studied cyber-attacks on availability, confidentiality, and integrity of data within MGs and also; it focused on FDIAs and presented several countermeasures. In [15], a cyber-attack detection model for stealthy cyber-attacks in MGs' control systems is proposed. The attack in that paper is performed through the man-in-the-middle scenario. The detection model uses machine learning algorithms to detect attacks. Reference [16] proposed a robust algorithm to detect FDIAs on DC MGs' current sensors. In that paper, first, the system is transformed to an observable state, and then the system states are estimated using a high-order sliding-mode observer.

[1]M. Mohammadi, A. Kavousifard, and M. Gitizadeh are with the Department of Electrical and Electronics Engineering, Shiraz University of Technology, Shiraz, Iran. (Email: Mojtabamohammadi303@gmail.com ,Gitizadeh@sutech.ac.ir, Kavousi@sutech.ac.ir). A. Aflaki is with the Schulich school of engineering university of Calgary, Calgary, Canada (Email: Arshia.aflaki@ucalgary.ca)

While each of the above works investigated a specific topic about cyber-security in smart grids, the research in this area is still in infancy. The research gap in cyber-attack detection within AMIs motivated us to write this paper. Among all types of attacks, the DIA is one of the most destructive and most common attacks. In this regard, this paper is mainly focused on DIAs. This paper proposes a novel cyber-attack detection algorithm based on an unsupervised learning method empowered by load forecasting methods to tackle the cybersecurity problem in AMIs. Note that previous works mainly use traditional methods such as Kalman Filters or upper and lower bound detection to detect cyber-attacks. The proposed method employs a forecasting model to predict the received load data and after preprocessing uses the Elliptic Envelope (EE) outlier detection method, which is an unsupervised learning method, to cluster the received data and detect attacks. In case of forecasting models, two different deep learning-based methods called BiLSTM and LSTM, along with one ensemble learning-based method called Extra Tree Regressor (ETR) are used and their results are compared. The real-time load data of the city of Johor in Malaysia is employed for training purposes and evaluating the proposed algorithm. Our detecting algorithm examines several DIA scenarios with different volumes of randomly picked data while they are increased or decreased in various levels. As it is illustrated in section VI, in most cases, the BiLSTM-EE detection method outperformed others. Some exceptions, however, occurred in a small number of scenarios. The main contribution of this work is as below:

- proposing a novel artificial-intelligence cyber-attack detection model for AMIs within smart grids.
- Comparing the results of three different forecasting methods (LSTM, BiLSTM, and ETR) considering different attack scenarios with different severities.

The rest of this paper is organized as follows: The basic theory of ETR, deep learning-based forecasting models (i.e., LSTM and BiLSTM), and EE is described in Sections II, III. The mathematical model of the proposed DIA detection method is presented in Section IV. Section V is devoted to simulation results and future works, and finally, the paper's main conclusion is presented in Section VI.

## II. LOAD FORECASTING MODELS

in this section, three load forecasting models are described and their mathematical model are presented.

### A. Extra Trees Regressor

ETR is a machine learning ensemble algorithm. It is a decision tree ensemble similar to other decision tree ensembles algorithms such as bootstrap aggregation (bagging) and random forest [17]. The Extra Trees algorithm uses the training dataset to generate a large number of unpruned decision trees. In the case of regression, predictions are made by averaging the prediction of the decision trees, while in the case of classification, majority voting is used. Unlike bagging and random forest, which construct each decision tree from a bootstrap sample of the training dataset, Extra Trees match each decision tree to the entire training dataset. Like the random forest, the Extra Trees algorithm will sample features at each split point of a decision tree at random. Unlike random forest, which chooses an optimal split point using a greedy algorithm, the Extra Trees algorithm chooses a split point at random.

The number of decision trees in the ensemble, the number of input features to pick and consider for each split point randomly, and the minimum number of samples needed in a node to generate a new split point are the three key hyperparameters to tune in the ETR algorithm. The algorithm's variance is increased by the random collection of split points, which makes the decision trees in the ensemble less correlated. By the number of trees in the ensemble, this increase in variance can be mitigated. Assume that $B$ is the number of training samples, $f$ is the forecasted value, and $x'$ is the non-tested sample, the training process of non-tested data is as follows [17].

$$f = \frac{1}{B}\sum_{b=1}^{B} f_b(x') \quad (1)$$

Where the standard deviation can be formulated as below:

$$\sigma = \sqrt{\frac{\sum_{b=1}^{B}(f_b(x') - f)^2}{B-1}} \quad (2)$$

### B. LSTM

The concept of LSTM, which is a recurrent neural network (RNN), is proposed to address and solve the gradient vanishing problem in conventional RNNs [18]. The LSTM uses a cell state and three cell gates (i.e., input gate, forget gate, and output gate) to maintain and adjust information in the data sequence.

The forget/input gate specifies the information that should be removed/added from/to the cell state, and the output gate specifies which information from the cell state should be used in the output. An illustration of the LSTM cell is presented in Fig. 1 where σ presents the logistic sigmoid function. It is worth noting that model's parameters (i.e., weights and biases) are trained through the backpropagation process [19].

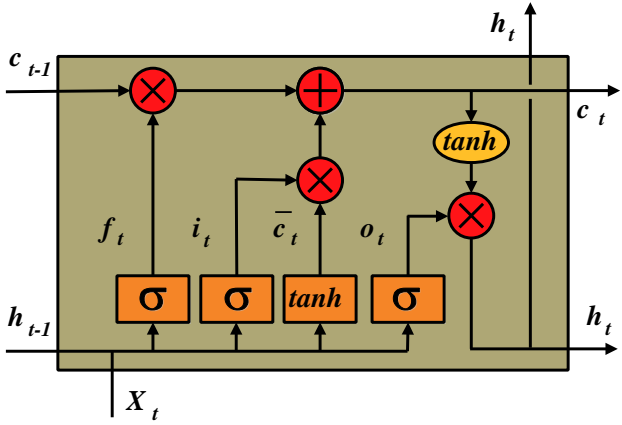

Fig. 1 structure of LSTM cell

In LSTM, the output sequence $(y_1, y_2, ..., y_T)$ is computed based on the input sequence $(x_1, x_2, ..., x_T)$ and hidden state sequence $(h_1, h_2, ..., h_T)$. The mathematical model of the LSTM cell is presented below.

$$h_t = H(W_{x,h}x_t + W_{h,h}h_{t-1} + b_h) \quad (3)$$

$$y_t = W_{h,y}h_t + b_y \quad (4)$$

The function H is implemented as follows:

$$i_t = \partial(W_{i,x}x_t + W_{i,h}h_{t-1} + b_i) \quad (5)$$

$$f_t = \partial(W_{f,x}x_t + W_{f,h}h_{t-1} + b_f) \quad (6)$$

$$\overline{c}_t = \tanh(W_{\overline{c},x}x_t + W_{\overline{c},x}h_{t-1} + b_{\overline{c}}) \quad (7)$$

$$o_t = \partial(W_{o,x}x_t + W_{o,h}h_{t-1} + b_o) \quad (8)$$

$$C_t = f_t \cdot c_{t-1} + i_t \cdot \overline{c}_t \quad (9)$$

$$h_t = o_t \cdot \tanh(c_t) \quad (10)$$

### C. BiLSTM

In BiLSTM, which is considered a modified version of LSTM and the forward layer, there is a backward layer that makes it possible for the model to use future contexts as well [20]. This feature of BiLSTM makes the model more potent in learning the behavior of data. In BiLSTMs, two layers operate in reversed time step directions such that the hidden forward sequence ($\vec{h}$) is generated by iterating the forward layer from $t=1$ to $t=T$. Similarly, backward hidden sequence ($\overleftarrow{h}$) is generated by iterating the backward layer from $t=T$ to $t=1$, and the and output sequence $(y)$ can be obtained using the following equations:

$$\vec{h}_t = H(W_{x,\vec{h}}x_t + W_{\vec{h},\vec{h}}\vec{h}_{t-1} + b_{\vec{h}}) \quad (11)$$

$$\overleftarrow{h}_t = H(W_{x,\overleftarrow{h}}x_t + W_{\overleftarrow{h},\overleftarrow{h}}\overleftarrow{h}_{t-1} + b_{\overleftarrow{h}}) \quad (12)$$

$$y_t = W_{\vec{h},y}\vec{h}_t + W_{\overleftarrow{h},y}\overleftarrow{h}_t + b_y \quad (13)$$

In the above formulas, $W_{f,x}$, $W_{i,x}$, $W_{o,h}$, ... present the weighting matrices, and $b_i$, $b_f$, $b_o$, ... are the bias vectors.

### III. ELLIPTIC ENVELOPE

In this section, we propose our outlier detection model for cyber-attack detection. EE, which is mainly utilized for outlier detection, finds a core for accurate data and also is able to separate data from polluting ones. The covariance EE can suit the data with a robust covariance calculation and an ellipse to the central data points ignores the points outside the central mode. The assumption that the standard data comes from a known distribution is a common way of doing outlier detection (e.g., data are Gaussian distributed). We try to describe the "form" of the data based on this assumption, and we can define outlying observations as observations that are sufficiently far from the fit shape. The model, for example, estimate the inlier position and covariance robustly if the inlier data are Gaussian distributed (i.e., without being influenced by outliers). The Mahalanobis distances derived from this calculation are used to calculate an outlining's metric. This technique is depicted in Fig. 2 and formulated as below [21].

$$\{d\}_{MH} = \sqrt{(X-\mu)^T C^{-1}(x-\mu)} \quad (14)$$

$x$ presents the data, μ is the mean of previous data, and $C^{-1}$ is the covariance matrix. P. C. Mahalanobis introduced the Mahalanobis distance in 1936 as a measure of the distance between a point P and a distribution D. [21] It is a multidimensional generalization of the concept of measuring how far P is from the mean of D. This distance is zero for P at the mean of D and increases as P moves away from the mean along each axis of the principal component. If each of these axes is rescaled to have unit variance, then the Mahalanobis distance in the transformed space corresponds to the standard Euclidean distance. Thus, the Mahalanobis distance is unitless, scale-invariant, and takes into account the data set's correlations.

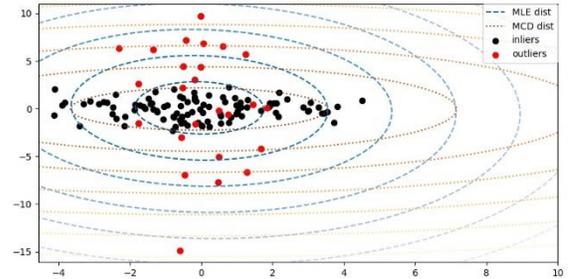

Fig. 2. Mahalanobis distances for a random dataset.

### IV. PROPOSED ARTIFICIAL INTELLIGENCE-BASED CYBER-ATTACK DETECTION MODEL

This section proposes our detecting algorithm with EE as outlier detection and load forecasting methods employed for data preprocessing. Electricity consumption forecasting data are often in large shape, so detecting their outliers by unsupervised learning methods is a challenging task to carry out. Preprocessing the data received by the SCADA system facilitates clustering algorithms for detecting the integration of false data in a dataset. Using the forecasted data, we can give each actual data sent for SCADA a rhythm making them easier to cluster. Let assume that the real load derived from AMIs at time $t$ is $y_t$ and the predicted load is $y_{pred,t}$. The preprocessing method can be formulated as below.

$$y_{processed,i} = \left| \frac{y_t - y_{pred,t}}{y_{pred,t}} \right| \quad (15)$$

In which $y_{processed,t}$ is the processed data. Assume that an attack vector sums with actual data and its components can be positive, negative, or a combination of both depending on the attack. Therefore, we can formulate attack as:

$$y_{processed,t,attacked} = \left| \frac{(y_t + a(t)) - y_{pred,t}}{y_{pred,t}} \right| \quad (16)$$

For anomaly detection, from (14), EE chooses an upper and lower band around $\{d\}_{MH}$ with respect to the mean values and covariances, meaning that every data with more or fewer distances than the mentioned bands will be detected as an anomaly. By combining (14) and (15), we can rewrite the EE formula as:

$$\{d\}_{MH} = \sqrt{\left(\left|\frac{(y_t - y_{pred,t})}{y_{pred,t}}\right| - \mu\right)^T C^{-1} \left(\left|\frac{(y_t - y_{pred,t})}{y_{pred,t}}\right| - \mu\right)} \quad (17)$$

With respect to historical data, EE will detect the attack and is able to upgrade the contamination, which is a key hyperparameter of EE in every hour. Let us combine (14) and (16).

$$\{d\}_{MH} = \sqrt{\left(\left|\frac{(y_t + a(t)) - y_{pred,t}}{y_{pred,t}}\right| - \mu\right)^T C^{-1} \left(\left|\frac{(y_i + a(t)) - y_{pred,t}}{y_{pred,t}}\right| - \mu\right)} \quad (18)$$

From (18), it is clear that by the minor difference between actual and predicted value, which leads to more minor mean and covariance, the outlier detection method will find attacked data with lower intensity easier. This highlights the vital role of accurate load forecasting in the detection algorithm.

Algorithm. 1 illustrates the procedure of the proposed method.

**Input**: Received load data
**Output**: A binary decision about cyber attack
1: Receive the actual data from meter at time t
2: Compute the corresponding forecasted data using forecasting model
3: Perform data processing (using equation (15))
4: Perform EE method (using equation (14))
5: **If** clustered as legitimate **then** add the data to the legitimate dataset
6: **if** clustered as outlier **then** Update contamination and replace the actual data with forecasted data
7: Wait for the next data and restart the process

Algorithm. 1. How the proposed method works.

Some evaluation metrics are used to check the robustness and accuracy of the cyber-attack detection method. In Table I, some basic measuring terms are described. The evaluation metrics used for this study are formulated as below.

$$Accuracy = \frac{TN + TP}{TN + TP + FN + FP} \quad (21)$$

$$Precision = \frac{TP}{TP + FP} \quad (22)$$

$$Sensitivity = \frac{TP}{TP + FN} \quad (23)$$

$$Specificity = \frac{TN}{TN + FP} \quad (24)$$

$$f_1 Score = \frac{2 \times Precision \times Sensitivity}{Precision + Sensitivity} \quad (25)$$

TABLE I
performance measuring terms

| Term | Description |
|---|---|
| True Positive (TP) | Attacked data correctly clustered as attacked |
| True Negative (TN) | legitimate data correctly clustered as not attacked |
| False Positive (FP) | Legitimate data incorrectly clustered as attacked |
| False Negative (FN) | Attacked data incorrectly clustered as not attacked |

V. NUMERIC RESULTS AND EXPERIMENTS

This section is devoted to the simulation results and comparison. In this section, first six regression models (i.e. Artificial Neural Network (ANN), BiLSTM, LSTM, Random Forest (RFR) [22], Gradient Tree Boosting (GTBR) [23], and Extra Tree Regression) are trained on the test case dataset and their results are compared. Then, the three best methods are selected and the EE is deployed on their output to test the proposed algorithm considering different forecasting models. To this end, two types of DIAs and several scenarios are considered to evaluate the performance of the proposed method under different cyber-attacks on AMI.

In this work, the hourly load data of the power company of the city of Johor in Malaysia is utilized to examine the performance of the proposed cyber-attack detection model. To this end, 15 months of data from 01/01/2009 to 14/03/2010 is used to train the model, and 3.5 months of data from 15/03/2010 to 02/07/2010 is used to evaluate the performance of the trained models. Also, two metrics called mean absolute percentage error (MAPE) and root mean square error (RMSE) are employed to evaluate the accuracy of proposed forecasting models. The ensemble learning methods are carried out using the Sklearn library [24], and deep learning methods are conducted by the Tensorflow library [25]. As mentioned, three best methods with respect to their evaluation metrics are selected as load forecasting methods for our proposed algorithm to detect DIA. Mentioned metrics are formulated as below.

$$MAPE = \frac{1}{n} \sum_{t=1}^{n} |\frac{x_t - x_{pred,t}}{x_t}| \qquad (19)$$

$$RMSE = \sqrt{\frac{1}{n} \sum_{t=1}^{n} \frac{(x_t - x_{pred,t})^2}{x_t}} \qquad (20)$$

While $x_{pred,t}$ is the predicted value, $x_t$ is the actual value, and $n$ is the number of test data samples.

### A. Experiments setup

Simulations are conducted using Python 3.8 on a personal laptop with the processor Intel(R) Core(TM) i7-10750H CPU @ 2.60GHz 2.59 GHz and 16GB of RAM. In order to make the load forecasting comparison more valid, all six mentioned forecasting methods are using the lookback of 14 hours loads. The LSTM and BiLSTM networks include two 128-cell layers, one dropout layer with 0.3 rate, and one fully connected layer. These models are trained considering 150 epochs, the Adam as the optimizer, and mean square error as loss function. The ensemble learning hyperparameters are optimized with Tensorflow Keras Optimizers [26], and the EE method is conducted by the Sklearn library.

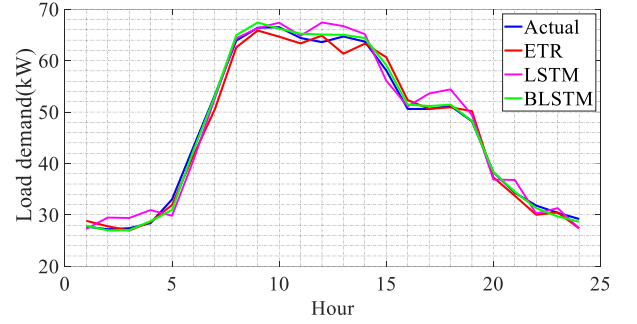

Fig. 1. The actual and predicted hourly loads of the test dataset.

### B. Benchmarking performance

In Table II, mentioned forecasting methods are compared with each other by their metrics scores. It is clear that BiLSTM and ETR are the most accurate forecasting models by about 2 percent MAPE error. LSTM is the second leading method for forecasting. The accuracy, however, decreases for RFR and GTBR, while the ANN scored the worst MAPE/RMSE error among all methods. Therefore, we chose the three best scoring load forecasting models (i.e., BiLSTM, LSTM, and ETR) for our attack detection algorithm. As can be seen, the error percentages of BiLSTM, LSTM, and ETR are close to each other. Fig. 3 shows predicted data by three mentioned best forecasting methods considering a 24-hour interval from 18/03/2010-00:00 to 19/03/2010-23:00. As can be seen from that figure, all three models have a good performance, although the BiLSTM has a slightly better performance in sharp points.

TABLE II
Evaluation metrics of hourly load forecast in test data

| Metrics \ methods | ETR | LSTM | BiLSTM | ANN | RFR | GTBR |
|---|---|---|---|---|---|---|
| MAPE% | 2.087 | 2.193 | 2.004 | 4.153 | 2.321 | 2.886 |
| RMSE(W) | 1336 | 1362 | 1322.4 | 2259 | 1583 | 1843 |

TABLE III
Different attack scenarios and their settings

| k% \ p% | -10 | -20 | -30 | -40 | -50 | +10 | +20 | +30 | +40 | +50 |
|---|---|---|---|---|---|---|---|---|---|---|
| 10 | Scenario 1 | Scenario 2 | Scenario 3 | Scenario 4 | Scenario 5 | Scenario 16 | Scenario 17 | Scenario 18 | Scenario 19 | Scenario 20 |
| 20 | Scenario 6 | Scenario 7 | Scenario 8 | Scenario 9 | Scenario 10 | Scenario 21 | Scenario 22 | Scenario 23 | Scenario 24 | Scenario 25 |
| 30 | Scenario 11 | Scenario 12 | Scenario 13 | Scenario 14 | Scenario 15 | Scenario 26 | Scenario 27 | Scenario 28 | Scenario 29 | Scenario 30 |

TABLE IV
Performance of proposed algorithms under attacks targeting blackouts

| Method \ Scenario | EE-ETR | | | EE-LSTM | | | EE-BiLSTM | | |
|---|---|---|---|---|---|---|---|---|---|
| | Accuracy | Specificity | $f_1$ Score | Accuracy | Specificity | $f_1$ Score | Accuracy | Specificity | $f_1$ Score |
| 1 | 95.48% | 97.48% | 78.06% | 95.56% | 97.52% | 78.44% | **96.17%** | **97.86%** | **81.41%** |
| 2 | 99.23% | 99.57% | 95.93% | **99.61%** | **99.78%** | **97.96%** | 99.30% | 99.61% | 96.34% |
| 3 | 99.77% | 99.87% | 98.96% | 99.84% | 99.91% | 99.31% | **99.84%** | **99.91%** | **99.31%** |
| 4 | **99.92%** | **99.95%** | **99.65%** | 99.92% | 99.95% | 99.65% | 99.92% | 99.95% | 99.65% |
| 5 | **100%** | **100%** | **100%** | 100% | 100% | 100% | 100% | 100% | 100% |
| 6 | 94.48% | 96.46% | 87.54% | 94.33% | 96.36% | 87.19% | **95.33%** | **97.00%** | **89.44%** |
| 7 | 99.23% | 99.51% | 98.15% | **99.46%** | **99.66%** | **98.70%** | 99.15% | 99.46% | 97.97% |
| 8 | **99.70%** | **99.82%** | **99.21%** | **99.70%** | **99.82%** | **99.21%** | **99.70%** | **99.82%** | **99.21%** |
| 9 | 99.92% | 99.95% | 99.65% | 99.92% | 99.95% | 99.65% | **100%** | **100%** | **100%** |
| 10 | **100%** | **100%** | **100%** | **100%** | **100%** | **100%** | **100%** | **100%** | **100%** |
| 11 | 93.03% | 94.98% | 88.59% | 92.57% | 94.65% | 87.84% | **94.25%** | **96.86%** | **90.06%** |
| 12 | 99.15% | 99.40% | 98.55% | **99.38%** | **99.56%** | **98.94%** | 99.15% | 99.40% | 98.55% |
| 13 | 99.69% | 99.78% | 99.47% | 99.77% | 99.83% | 99.60% | **99.84%** | **99.89%** | **99.73%** |
| 14 | 99.92% | 99.94% | 99.87% | 99.92% | 99.94% | 99.87% | **100%** | **100%** | **100%** |
| 15 | **100%** | **100%** | **100%** | **100%** | **100%** | **100%** | **100%** | **100%** | **100%** |

### C. Cyber-Attacks settings and scenarios

By attacking the data derived from AMI, power system operators can be easily fooled by the false data leading them to make wrong decisions. We introduce two common DIAs in this section where one of them decreases the actual value of data received by SACADA, while the other increases them. Assume that the hackers randomly target about $k$ percent of our data and either decrease or increase them by $p$ percent, so the $p$ parameter represents attack intensity and $k$ denotes dispersion of the attack. Using three different $k$, from 10 to 30, and various $p$, ranging from 10 to 50, thirty scenarios are created. All of attack scenarios and their settings are illustrated in Table III. It is worth noting that we added some Gaussian noises to the non-attacked data with the severity of 2 percent, as our forecasting methods are predicting the loads with about 98 percent accuracy, to make the scenarios more realistic.

Generally, by decreasing the actual data, attackers target blackouts and other system failures in the power system. Decreasing the load data derived from buses will manipulate the generation sector to decrease their power generation, so the electricity generation fails to meet demand. In the worst-case scenario, when some of the utilities have ramp rates or generation limits in islanded grids, keeping up with demand is not possible in few seconds. This insufficient capacity expansion in the generation sector decreases reliability indices and increases the risk for blackouts. On the other hand, increasing the actual data of loads fools the operators and leads the utilities to generate more than demanded power. This overgeneration may cost power companies unnecessary expenses and force them to use their spinning reserves or peak units, leading to economic loss. Therefore, it is logical to divide the attack scenarios into two portions: targeting blackouts and targeting economic loss.

### D. Performance evaluation and discussion

By using the two types of cyber-attacks that are mentioned in the last subsection, our proposed detection algorithms are tested under several scenarios introduced in Table III.

TABLE V
Performance of proposed algorithms under attacks targeting economic loss

| Method<br>Scenario | EE-ETR | | | EE-LSTM | | | EE-BiLSTM | | |
|---|---|---|---|---|---|---|---|---|---|
| | Accuracy | Specificity | $f_1$ Score | Accuracy | Specificity | $f_1$ Score | Accuracy | Specificity | $f_1$ Score |
| 16 | 96.09% | 97.83% | 79.84% | 95.94% | 97.75% | 79.05% | **96.70%** | **98.17%** | **83.00%** |
| 17 | 99.15% | 99.52% | 95.98% | **99.31%** | **99.61%** | **96.71%** | 99.15% | 99.52% | 95.98% |
| 18 | 99.61% | 99.78% | 98.13% | **99.77%** | **99.87%** | **98.88%** | **99.77%** | **99.87%** | **98.88%** |
| 19 | 99.84% | 99.91% | 99.15% | 99.92% | 99.95% | 99.57% | **100%** | **100%** | **100%** |
| 20 | 99.92% | 99.95% | 99.59% | **100%** | **100%** | **100%** | **100%** | **100%** | **100%** |
| 21 | 94.18% | 96.34% | 85.79% | 94.10% | 96.29% | 85.60% | **95.40%** | **97.11%** | **90.01%** |
| 22 | 98.92% | 99.32% | 97.33% | 98.92% | 99.32% | 97.33% | **99.08%** | **99.42%** | **97.71%** |
| 23 | 99.61% | 99.76% | 99.05% | 99.77% | 99.85% | 99.43% | **99.84%** | **99.90%** | **99.62%** |
| 24 | 99.77% | 99.85% | 99.40% | **99.84%** | **99.90%** | **99.60%** | **99.84%** | **99.90%** | **99.60%** |
| 25 | 99.92% | 99.95% | 99.81% | 99.92% | 99.95% | 99.81% | **100%** | **100%** | **100%** |
| 26 | 92.42% | 94.63% | 87.12% | 92.49% | 94.68% | 87.25% | **94.64%** | **96.20%** | **90.89%** |
| 27 | 98.85% | 99.17% | 98.12% | **99.00%** | **99.28%** | **98.37%** | **99.00%** | **99.28%** | **98.37%** |
| 28 | 99.69% | 99.78% | 99.49% | **99.77%** | **99.83%** | **99.62%** | 99.54% | 99.67% | 99.24% |
| 29 | 99.84% | 99.88% | 99.75% | **99.92%** | **99.94%** | **99.87%** | 99.84% | 99.88% | 99.75% |
| 30 | 99.84% | 99.88% | 99.75% | 99.92% | 99.94% | 99.87% | **100%** | **100%** | **100%** |

Table IV illustrates the numeric results of the proposed method under fifteen DIA scenarios targeting blackouts. It is worth noting that the algorithms are evaluated by three statistical metrics introduced in this paper. In table V and IV, leading percentages are bolded, intermediate percentages, between 89 and 80, are shaded in yellow, and not acceptable percentages are shaded in red. For our evaluation metrics, percentages that are more than 90 are assumed to be excellent. It is worth noting that all the proposed models are tested three times under different scenarios and the mean value of each metrics is wrote in the respective tables. From Table IV, it is clear that the EE-BiLSTM detection algorithm outperformed other two algorithms. In scenarios with lowest intensity attacks, scenarios 1, 6 and 11, both EE-ETR and EE-LSTM incorrectly clustered some actual data as attacked and vice versa, which decreased their lowest $f_1$ Score to below 80 percent. EE-BiLSTM method managed to limit its lowest $f_1$ Score to 81 percent, which is assumed to be an intermediate performance with the number of FPs and FNs. By increasing the $k$, the number of FPs and FNs reduced significantly, meaning that all proposed algorithms are able to detect attacks with higher scattering. When the attack severity escalated, metrics were boosted for all detection algorithms in such a way that by reaching the highest intensity, methods were able to detect the attacked data correctly independent of $k$. The accuracy and specificity of all algorithms never fell under 92 and 94 percent, respectively. EE-BiLSTM scores better than the other ones in most scenarios making the mentioned method the most accurate and robust detection algorithm. It is worth noting that as the accuracies of the proposed forecasting methods are significantly high, our proposed cyber-attack detection method was able to detect a great portion of the cyber-sabotages.

It is observable from Table V that in most scenarios, the performance of EE-BiLSTM method is better than other detection methods. Similar to the previous DIA, when $p$ was in its lowest value, attack detection conducted by both EE-ETR and EE-LSTM failed to limit the number of FPs and FNs led to not acceptable $f_1$ Scores in one scenario and two $f_1$ Scores that were less than 90 percent in two scenarios. EE-BiLSTM limited the numbers of its false positives and false negatives, leading to only one scenario with $f_1$ Score lower than 90 percent. In DIA targeting economic loss, only EE-BiLSTM managed to reach zero FPs and FNs in more than one scenario. By increasing the $k$, the number of FPs and FNs reduces significantly in lower intensity cases, which means that the $f_1$ Score is rising. In Table VI, the number of scenarios in which the proposed detection methods had leading numbers is illustrated with their rankings.

*E. Other cyber-attacks and future works*

Although we only considered DIAs, which is one of the most common data attacks, there are other types of attacks that can be studied. Denial of Service (DoS) is one of the famous cyber-sabotages, but the attack is easily detectable for operators due to the nature of data received by SCADA.

Some numerous algorithms and methods can be employed for cyber-attack detection. It is worth noting that numerous load forecasting models are being generated nowadays, so it is possible to use more accurate and robust forecasting methods to increase the percentages of the proposed indices in this paper. Kalman filters and other methods that using forecasting threshold to detect polluted data are studied thoroughly. Hence, the mentioned methods are able to collaborate with unsupervised learning techniques for developing novel architecture algorithms.

In this study, we did not calculate the economic loss or blackout probability. As future work, it is possible for some datasets to be used as real loads in different microgrids to study the effect of DIA on power grids from economic and probability perspectives. At the end, we strongly encourage researchers to use other outlier detection methods, such as isolation forest, with hyperparameters optimizing for cyber-attack detection.

TABLE VI
Ranking of the detection models with respect to leading scenarios numbers

| Methods | Number of leading scenarios | Ranking |
|---|---|---|
| EE-ETR | 5 | 3 |
| EE-LSTM | 15 | 2 |
| EE-BiLSTM | 24 | 1 |

VI. CONCLUSION

This paper proposed an artificial intelligence-based cyber-attack detection model for DIAs on load data in smart grids. The proposed model uses an artificial intelligence-based regression algorithm to predict the next load data and uses an unsupervised learning algorithm called elliptic envelope to cluster the preprocessed received data and label the data (attack or no attack). The forecasting methods in this work consist of two deep-learning algorithms called LSTM and BiLSTM, along with one ensemble learning algorithm called ETR. The proposed model was evaluated using a practical data set related to the power company of the city of Johor in Malaysia. The results showed that Bilstm has the best performance in load forecasting, although the other two algorithms (i.e., LSTM and ETR) are acceptable. In the case of EE-BiLSTM, evaluating scenarios demonstrated that the method is able to (i) accurately detect more than 94 percent of the outliers and label them as attacked data even when the severity of attacks were low, (ii) decreases the number of FP and FN comparing to other proposed methods, and (iii) is not affected by widely scattered set of attacked data. The other two methods failed to decrease their FPs and NPs in scenarios with the lowest attack intensity. However, their accuracy and specificity were near to the EE-BiLSTM method.